\documentclass{article}

\usepackage[margin=1in]{geometry}
\usepackage[table]{xcolor}

\usepackage{graphicx}
\usepackage{booktabs}
\usepackage{amsmath}
\usepackage{amssymb}
\usepackage{amsfonts}
\usepackage{multirow}
\usepackage{verbatim}
\usepackage{caption}
\usepackage{longtable}
\usepackage{subfig}
\usepackage{supertabular}
\usepackage{float}
\usepackage{lscape}
\usepackage{fancyhdr}

\usepackage{tablefootnote}
\usepackage[round,semicolon]{natbib}
\usepackage{xspace}
\usepackage{textcomp}
\usepackage{makecell}
\usepackage{siunitx}
\usepackage{colortbl}

\usepackage{pifont}
\definecolor{darkerGreen}{RGB}{0,170,0}

\usepackage[]{mdframed}

\usepackage[subfigure]{tocloft}

\usepackage{scrextend}

\usepackage{array}
\usepackage{tgpagella}
\usepackage{latexsym}
\usepackage[T1]{fontenc}
\usepackage{microtype}
\definecolor{mydarkblue}{rgb}{0,0.08,0.45}
\PassOptionsToPackage{hyphens}{url}\usepackage[colorlinks,citecolor=mydarkblue,urlcolor=mydarkblue,linkcolor=red,linktocpage=true]{hyperref}
\usepackage{url}            
\usepackage{nicefrac}       
\usepackage{changepage}
\usepackage{xargs}          
\usepackage{wrapfig,lipsum,booktabs}
\usepackage{endnotes}
\usepackage{adjustbox}

\usepackage{pgfplots}
\usetikzlibrary{pgfplots.groupplots}
\pgfplotsset{compat=1.3}
\usepackage{tikz}
\usetikzlibrary{patterns}

\usepackage[most]{tcolorbox}

\usepackage[capitalize,noabbrev]{cleveref}
\crefname{section}{Section}{\S\S}
\Crefname{section}{Section}{\S\S}
\crefname{table}{Table}{Tables}
\crefname{figure}{Figure}{Figures}
\crefname{algorithm}{Algorithm}{}
\crefname{equation}{eq.}{}
\crefname{appendix}{Appendix}{}
\crefformat{section}{Section #2#1#3}
\usepackage{multicol}

\usepackage{algorithm}
\usepackage{algorithmic}

\usepackage{xspace}
\usepackage[textsize=tiny]{todonotes}

\newlength{\myMheight}
\usepackage{minitoc}

\usepackage[bottom,symbol]{footmisc}

\vspace{-10mm}
\title{
    \textbf{Accelerating Flood Warnings by 10 Hours: The Power of River Network Topology in AI-enhanced Flood Forecasting
    }
}
\vspace{-3mm}

\renewcommand\footnotemark{}
\newcommand\tinyspace{\hspace{0.12em}}

\author{
    \normalsize{}
    \textbf{Hongjun Wang \tinyspace$^{\mathbf{1}}$}
    \hspace{4mm}
     \textbf{Jiyuan Chen  \tinyspace$^{\mathbf{2}}$}
    \hspace{4mm}
	\textbf{Yinqiang Zheng \tinyspace$^{\mathbf{1}}$}
	\hspace{4mm}
	\textbf{Xuan Song \tinyspace$^{\mathbf{3}}$}
	\\
	\\
	\normalsize{}
	$^{1}$ The University of Tokyo
	\hspace{4mm}
	$^{2}$ The Hong Kong Polytechnic University
	\hspace{4mm}
	$^{3}$ Jilin University
	\hspace{4mm}
	\\
	\\
	\normalsize{}
	Corresponding authors: Yinqiang Zheng (yqzheng@ai.u-tokyo.ac.jp), Xuan Song (songxuan@jlu.edu.cn)
	}

\newcommand{\eqnref}[1]{Eq.~(\ref{#1})}

\newcommand{\fref}[1]{Figure~\ref{#1}} 

\definecolor{ncred}{HTML}{e04c71}
\definecolor{unspecgold}{HTML}{e0cd92}
\definecolor{cblue}{HTML}{82b5cf}

\date{}
\begin{document}
\maketitle



\vspace{-1mm}
\begin{abstract}
\noindent
Climate change-driven floods demand advanced forecasting models, yet Graph Neural Networks (GNNs) underutilize river network topology due to tree-like structures causing over-squashing from high node resistance distances. This study identifies this limitation and introduces a reachability-based graph transformation to densify topological connections, reducing resistance distances. Empirical tests show transformed-GNNs outperform EA-LSTM in extreme flood prediction, achieving 24-hour water level accuracy equivalent to EA-LSTM’s 14-hour forecasts—a 71\% improvement in long-term predictive horizon. The dense graph retains flow dynamics across hierarchical river branches, enabling GNNs to capture distal node interactions critical for rare flood events. This topological innovation bridges the gap between river network structure and GNN modeling, offering a scalable framework for early warning systems.
\end{abstract}

\vspace{-2mm}
\section{Introduction}
\label{introduction}
\vspace{-1mm}
 \textcolor{black}{The global hydrological cycle is undergoing significant anthropogenic alterations, with far-reaching implications for flood risk and water resource management \cite{boulange2021role,nearing2024global,wing2022inequitable,shu2023integrating}. \citep{Milly2002} and \citep{Prein2016} highlighted the increasing risk of extreme precipitation events due to climate change, while \citep{Hirabayashi2013} and \citep{Winsemius2015} projected a substantial increase in global flood risk under future climate scenarios.}

 \textcolor{black}{The increasing frequency and complexity of flood events has spurblack significant developments in flood forecasting methodologies. Traditional approaches integrate meteorological inputs with hydrological processes to simulate rainfall-runoff dynamics and channel flows \citep{bartholmes2005coupling}. However, uncertainties in precipitation forecasts often limit the accuracy of these models \citep{speight2021operational,krzysztofowicz2001forecastuncertainty}. Recent advances in machine learning have introduced new possibilities, with LSTM networks demonstrating particular promise in capturing both linear and non-linear memory effects in hydrological time series modeling \citep{le2019application}. Convolutional Neural Networks (CNNs), though originally developed for image processing, have also proven effective in pblackicting hydrological time series \citep{wang2017deep,shi2015convolutional}.}


 \textcolor{black}{Feature selection plays a crucial role in improving model performance, with recent work by \cite{sheikhpour2025sparse} introducing innovative approaches using hypergraph Laplacian-based semi-supervised discriminant analysis. Their method effectively captures higher-order relationships while maintaining computational efficiency, particularly valuable for complex hydrological systems. Additionally, advances in flow optimization, such as the work by \cite{abdollahi2021flow} combining particle swarm optimization with knapsack algorithms, have important implications for modeling water flow dynamics in river networks.}

 \textcolor{black}{Pblackicting streamflow in ungauged basins continues to pose significant challenges, particularly for infrastructure projects related to flood management \citep{Grill2019,Abbott2019,plate2002flood}, hydropower generation \citep{robinson1997climate}, and agricultural water resources \citep{yang2021hydrological}. In response to this pressing need, the International Association of Hydrological Sciences (IAHS) launched the "Pblackiction in Ungauged Basins" (PUBs) initiative, which ran from 2003 to 2013 \citep{sivapalan2003iahs}, aiming to advance methodologies for streamflow forecasting in ungauged catchments. Researchers have developed and tested a range of statistical and physical models \citep{castiglioni2011smooth, skoien2007spatiotemporal, farmer2016ordinary, wagener2004rainfall, wagener2006parameter}, although substantial room for improvement remains.}

Recent advancements in streamflow simulation and regionalization have been notably driven by LSTM-based models \citep{hochreiter1997long}, which excel at capturing temporal dependencies. Among these, the Entity-Aware LSTM (EA-LSTM) has shown exceptional performance by integrating both meteorological and catchment data \citep{kratzert2018rainfall, kratzert2019toward}. Despite these strides, limitations persist, particularly in accounting for river network topology, as highlighted in recent studies \citep{kratzert2021large}. This shortcoming is partly attributed to the constraints of benchmark datasets such as CAMELS-x \citep{addor2017camels}, which do not inherently represent the hydrological connectivity between upstream and downstream locations.

While Graph Neural Networks (GNNs) offer promising solutions for modeling network structures, they often encounter challenges with over-squashing when dealing with large-scale or complex graphs. This phenomenon, where information from distant nodes becomes excessively compressed during message passing, has been extensively studied. Recent work has proposed various solutions, including graph rewiring methods based on curvature \citep{topping2021understanding}, spectral expansion theory \citep{karhadkar2022fosr}, and attention mechanisms \citep{wu2021representing,ying2021transformers}. These advances in addressing over-squashing are particularly relevant for hydrological applications, where preserving information flow across river networks is crucial.

The introduction of the LamaH-CE dataset \citep{klingler2021lamah}, which incorporates topological data, offers new research opportunities to address this gap. However, early studies \citep{kirschstein2024merit} have revealed that even with the application of graph neural networks to river network graphs, the influence of topology on pblackictive performance remains surprisingly insignificant. This counterintuitive finding, which suggests that river network graphs may offer no performance advantage over simple multilayer perceptrons (MLPs), raises important questions about the utility of topological information in flood forecasting models.

In \citep{kirschstein2024merit}, GNN architectures, including GCN \citep{kipf2017semi}, GAT \citep{velickovic2017graph}, and GCNII \citep{chen2020simple}, were employed for flood forecasting using the LamaH-CE dataset. By experimenting with different adjacency matrices, the study assessed the impact of varying topological definitions on model performance. Results showed that even when edges were entirely removed, the GNN performed comparably to an MLPs, with no tangible benefit from the inclusion of weighted edges that represent physical hydrological relationships. Given the inherent spatial correlations between upstream and downstream stations due to the fluid nature of water, this finding seems counterintuitive. The study in \citep{kirschstein2024merit}, however, does not fully explain the underlying causes of this topological ineffectiveness, prompting further exploration in this paper.

In this work, we investigate the underperformance of river network topology in GNNs, specifically from the perspective of over-squashing \cite{black2023understanding}, a phenomenon where distant nodes' signals become too compressed as they traverse the graph. We propose a simple yet physically motivated solution to mitigate over-squashing, and our empirical results reveal that the role of river network topology has been underestimated, particularly for long-term forecasting and the pblackiction of rare, large spike floods.   \textcolor{black}{While previous approaches have attempted to incorporate river network topology into flood forecasting models, they have been limited by the inherent constraints of tree-like structures in information propagation. Our novel dense graph transformation approach fundamentally reimagines how we represent river networks in GNNs, enabling more effective capture of both local and long-range dependencies in water flow dynamics.} 
These extreme flood events, often driven by the rapid convergence of nearby rivers, represent a case where GNNs' potential can be fully realized. We hope our findings will encourage further exploration of graph-based approaches in the development of more effective flood warning systems.


\begin{figure}[h]
	\centering
	\includegraphics[width=\linewidth]{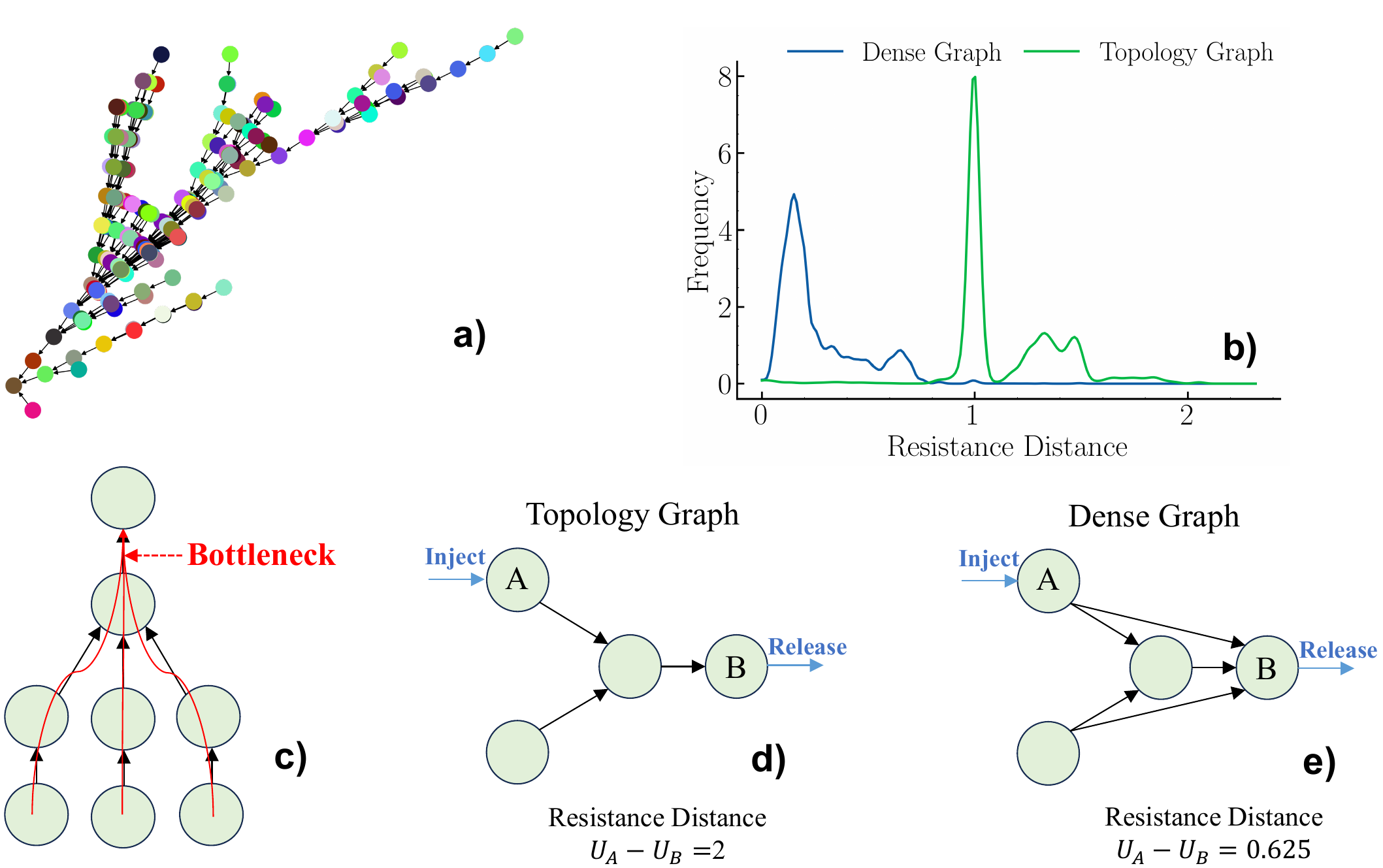}
	\caption{ \textcolor{black}{\textbf{Visual exploration and quantitative analysis of river network topology.} \textbf{a)} Visualization of an actual river network topology, showing the natural tree-like structure with multiple tributaries converging into main streams. \textbf{b)} Comparison of resistance distance distributions between dense and topology graphs, where the dense graph (blue line) shows lower resistance distances overall compablack to the topology graph (green line), indicating better information flow. \textbf{c)} Conceptual illustration of the bottleneck problem in tree-like structures, where information must pass through critical nodes that can become congested. \textbf{d,e)} Quantifying resistance distance through electrical analogy: When injecting 1A current at node A and extracting it at node B, the measublack voltage difference $V(A)-V(B)$ represents the resistance distance. The dense graph structure (e) blackuces this voltage difference from 2V to 0.625V compablack to the topology graph (d), demonstrating improved connectivity.}}
	\label{fig:graph}
\end{figure}

\begin{figure}[t]
	\centering
	\includegraphics[width=1\linewidth]{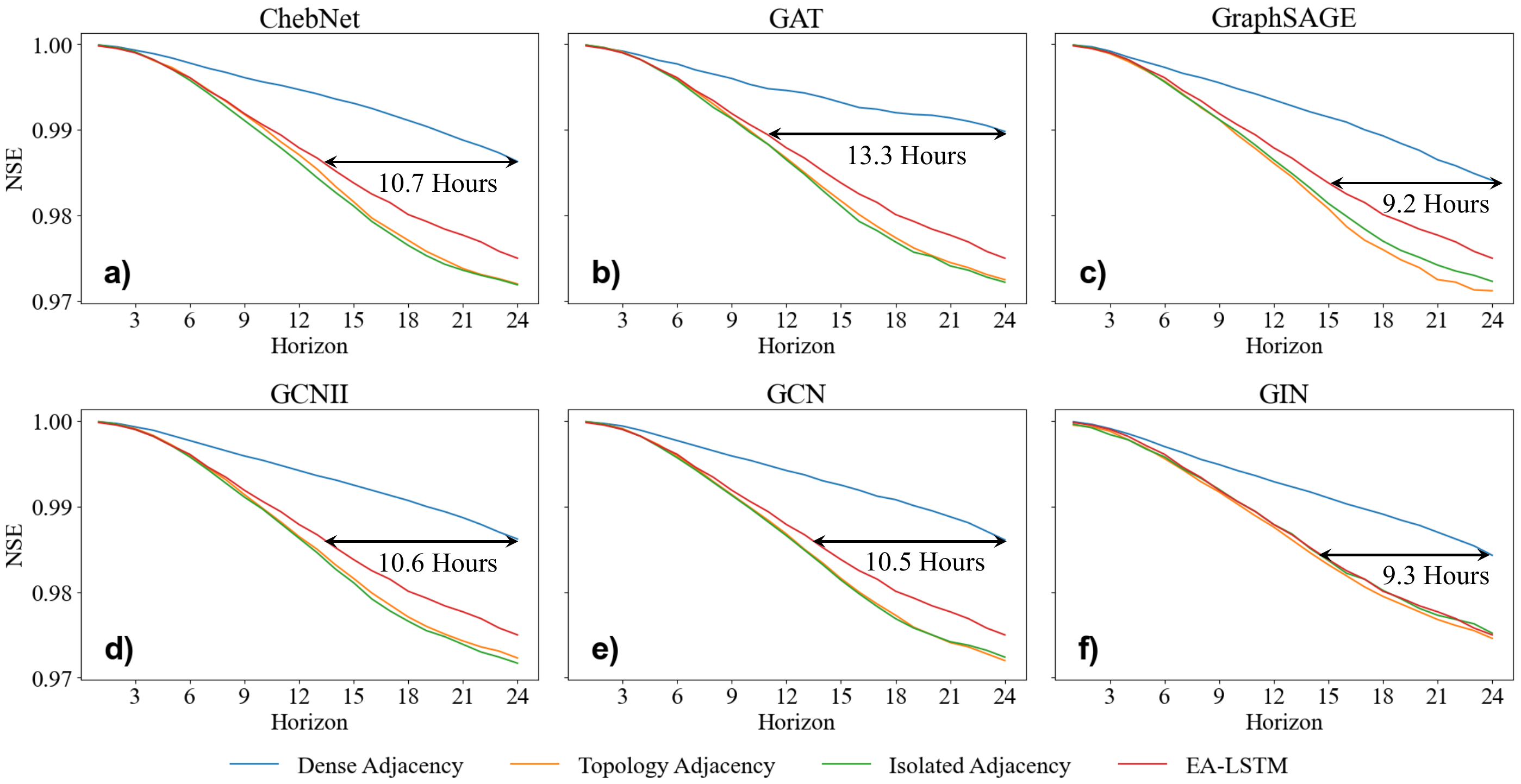}
	\caption{\textbf{In this study, we evaluate the performance of mainstream GNNs  by analyzing their Nash-Sutcliffe efficiency across different types of adjacency matrices (dense, topological, and isolated) over varying pblackiction horizons.} Additionally, the EA-LSTM model is employed as a baseline for comparison, allowing us to emphasize the superior long-term forecasting capabilities of GNNs relative to EA-LSTM. }
	\label{fig:nse}
\end{figure}



\begin{figure}[h]
	\centering
	\subfloat[Dense adjacency versus Isolated adjacency ]{\includegraphics[width=0.82\linewidth]{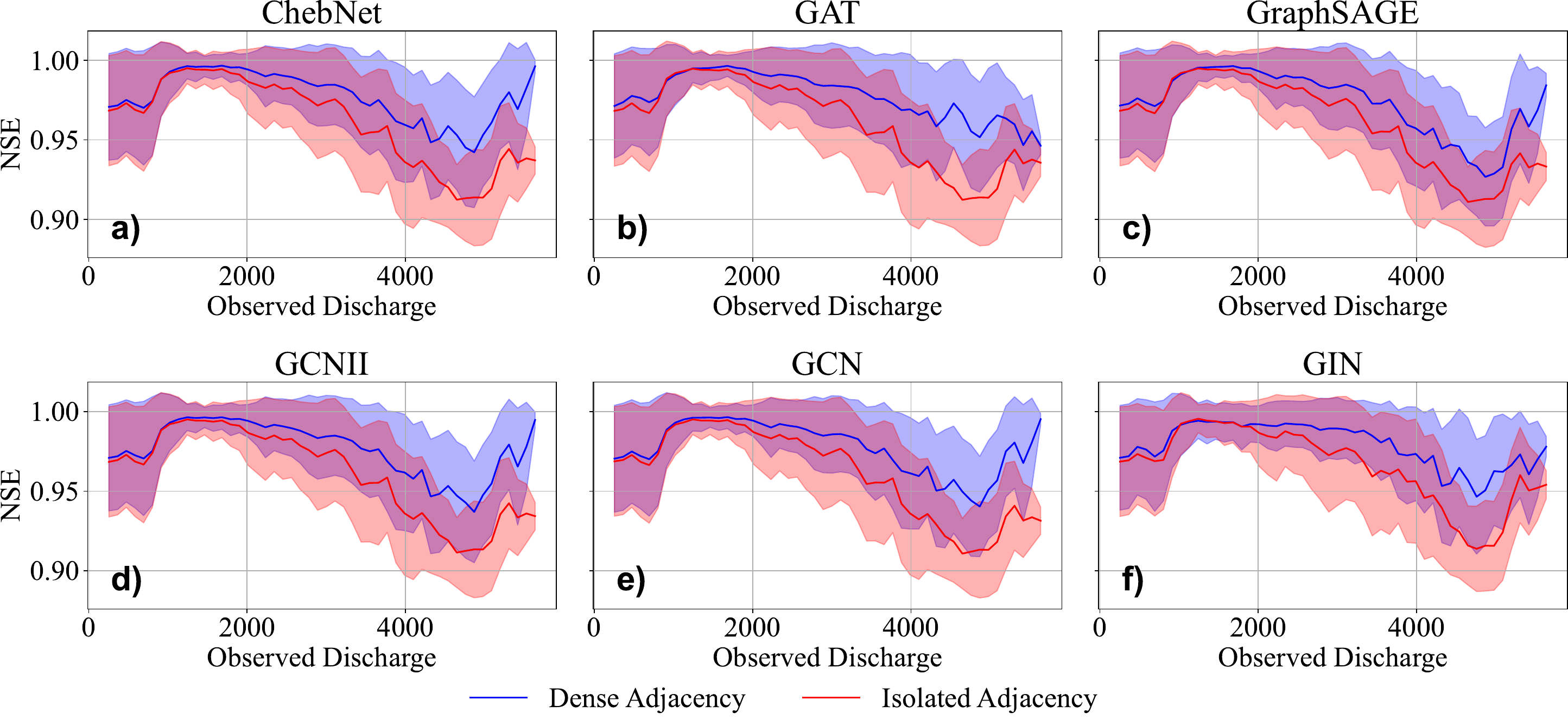}\label{fig:nse_dist_mlp}} \\
	\subfloat[Dense adjacency versus Isolated adjacency ]{\includegraphics[width=0.82\linewidth]{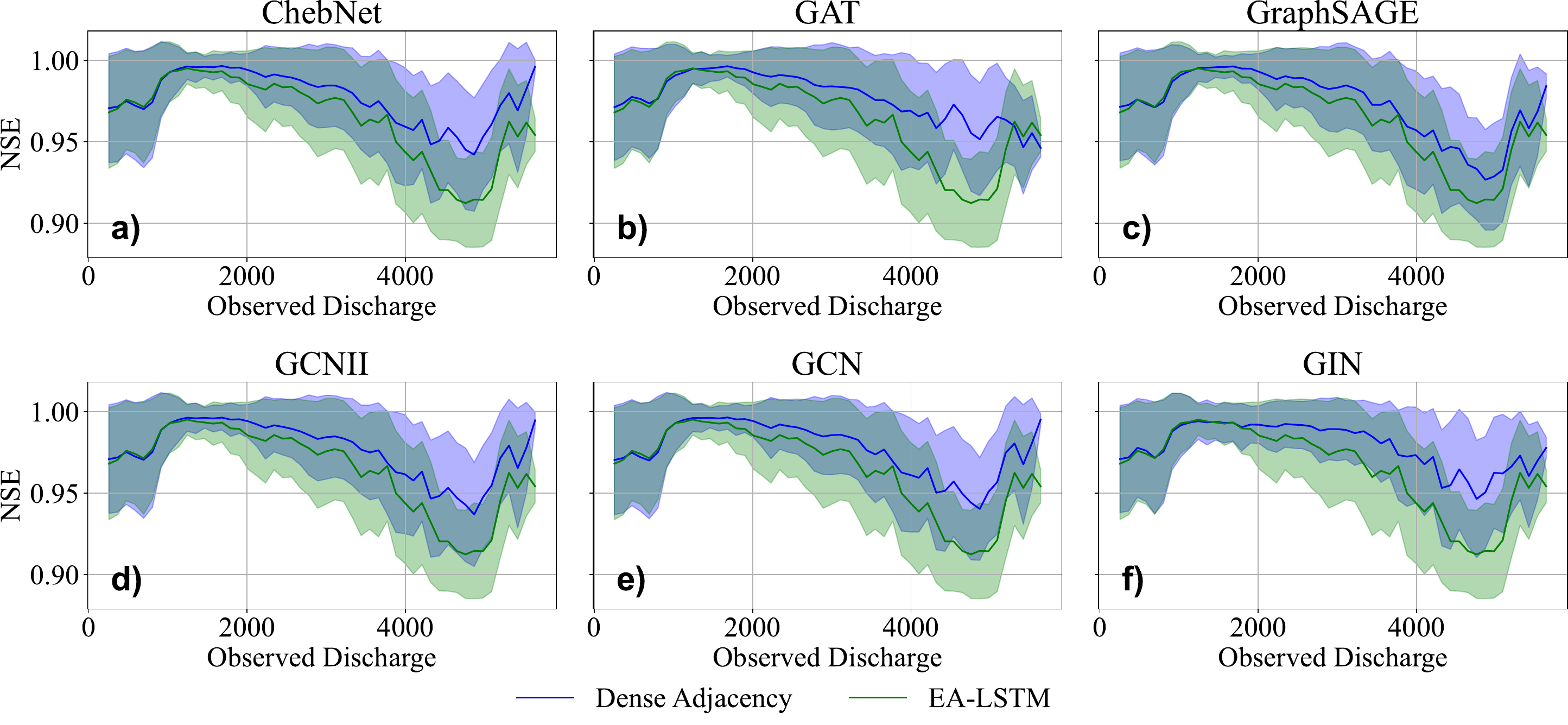}\label{fig:nse_dist_lstm}}	
	\caption{\textbf{\fref{fig:nse_dist_mlp} and \ref{fig:nse_dist_lstm} illustrate the performance of six different GNNs on LamaH-CE.} We compares the NSE trends of the models under three scenarios: dense adjacency, EA-LSTM, and isolated adjacency.   Overall, the NSE  across different discharge,  dense adjacency significantly outperforming both isolated adjacency and EA-LSTM. }
\end{figure}

\section{Results}
\paragraph{Problem Formulation.}
Let \( \mathcal{G} = (V, E, A) \) represent a stream network, where \( V \) and \( E \) denote the sets of nodes (gauges) and edges (flow directions), respectively. The matrix \( A \) represents the adjacency matrix of the stream network \( \mathcal{G} \). We define the gauge signal matrix \( X_{(t)} \in \mathbb{R}^{N \times C} \) for \( \mathcal{G} \), where \( C \) represents the dimensionality of the features, \( N = |V| \) is the number of vertices, and \( X_{(t)} \) denotes the observations of the spatial network \( \mathcal{G} \) at time step \( t \).
The flood forecasting task aims to learn a multi-step pblackiction function \( f \) based on past observations:
\[
f((X_{(t-\alpha)}, X_{(t-\alpha+1)}, \ldots, X_{(t-1)}), \mathcal{G}) \rightarrow (X_{(t)}, X_{(t+1)}, \ldots, X_{(t+\beta)})
\]
where \( \alpha \) represents the input length of past time step observations, and \( \beta \) denotes the number of future steps to be pblackicted.

 \textcolor{black}{\textbf{Analysis the Failure of River Topological in GNNs.}
The topological structure of river networks is a core subject of study in hydromorphology. This structure can be precisely described as an acyclic directed graph, specifically a spanning tree \citep{rodriguez1997fractal,rinaldo2014evolution}. As illustrated in Figure 1, river networks exhibit a distinctive tree-like structure, where each node (except the outlet) has exactly one downstream connection, while potentially having multiple upstream connections \citep{dodds2000scaling}. In Figure 2 a, we elaborate on the topological structure of the LamaH-CE network. \\
The hierarchical structure of river networks bears significant resemblance to the over-squashing issue \citep{topping2021understanding,alon2020bottleneck} observed in GNNs. Over-squashing is a key theoretical and practical challenge in GNNs, referring to the phenomenon where information from distant nodes becomes excessively compressed and distorted during message passing. This issue stems from the bottlenecks in the graph's topology, which hinder the flow of information between distant nodes. Theoretically, over-squashing can be quantified using the Resistance Distance \cite{black2023understanding}, a metric in graph theory that quantifies the relationship between nodes by modeling the graph as an electrical network, calculating the equivalent resistance between nodes to reflect the overall structure and connectivity.  Figure 2b compares the distribution characteristics of resistance distances between dense and topological graphs. The blue line represents the dense graph, showing a higher frequency at smaller resistance distances, while the green line represents the topological graph, exhibiting a prominent peak around a resistance distance of 1. The inset on the right uses a simplified network structure to illustrate how the graph topology influences the distribution of resistance distances. We highlight the concept of a "bottleneck," where central nodes in the topological graph create bottlenecks, whereas the dense graph displays a more uniform connectivity, leading to generally lower resistance distances.\\
Mathematically, the effective resistance between two nodes $u$ and $v$ can be expressed as:
\begin{equation}
	R_{u,v} = (1_u - 1_v)^T L^+(1_u - 1_v)
\end{equation}
where $L^+$ is the pseudoinverse of the graph Laplacian matrix, and $1_u$, $1_v$ are indicator vectors for nodes $u$ and $v$ respectively. In river networks, this resistance distance becomes particularly significant at confluence points, where information from multiple upstream tributaries must be compressed into fixed-dimensional node vectors. The tree-like structure inherently results in high resistance distances between nodes in different branches, directly impacting the ability of GNNs to propagate information effectively.\\
The impact on GNN performance can be quantified through the bound on the Jacobian matrix of node features:
\begin{equation}
	\frac{\partial h^{(r)}_u}{\partial x_v} \leq (2\alpha\beta)^r \frac{d_{max}}{2} \left(\frac{2}{d_{min}}\right) \left(\frac{r + 1 + \mu^{r+1}}{1 - \mu} - R_{u,v}\right)
\end{equation}
where $r$ is the number of layers, $\alpha$ and $\beta$ are model parameters, and $\mu$ is related to the eigenvalues of the normalized adjacency matrix. This bound demonstrates how higher resistance distances ($R_{u,v}$) directly limit the influence of upstream features on downstream pblackictions.\\
From a physical perspective, this limitation is particularly problematic for flood forecasting, where accurate pblackictions require preserving detailed information about upstream conditions and their complex interactions. During flood events, the rapid convergence of water from multiple tributaries creates complex dynamics that need to be captublack in the model. However, the tree-like structure's inherent high resistance distances make it difficult for GNNs to maintain and propagate this crucial information effectively.\\
To address these limitations, we propose transforming the original tree-like topology into a dense graph structure, as illustrated in Figure 2d,e. This transformation blackuces effective resistance between nodes while preserving the physical meaning of river network relationships. The dense graph enables more direct information flow between hydrologically connected locations, allowing the model to better capture both local and long-range dependencies in water flow dynamics.}

\paragraph{Mitigating Over-squashing with Dense Graph.}
\fref{fig:nse} presents a comparative analysis of six prominent Graph Neural Network models—ChebNet \citep{defferrard2016convolutional}, GAT \citep{velickovic2017graph}, GraphSAGE \citep{hamilton2017inductive}, GCNII \citep{chen2020simple}, GCN \citep{kipf2017semi}, and GIN \citep{xu2018powerful}—evaluating their performance in terms of Nash-Sutcliffe efficiency (NSE) across varying pblackiction horizons (1-24 hours), adjacency matrix types (dense, topological, and isolated) and baseline: EA-LSTM \cite{kratzert2018rainfall}. The results, displayed in six subplots (a-f), demonstrate a consistent trend of decreasing NSE as the pblackiction horizon extends, with dense adjacency matrices consistently outperforming their topological, isolated counterparts and EA-LSTM, particularly in long-term pblackictions. The performance in \fref{fig:nse} disparity becomes even more pronounced in long-term forecasting, underscoring the significant advantage of graph-based information for extended flood pblackiction horizons.  We present a comparison of the leading time in GNNs versus EA-LSTM pblackictions. Specifically, we evaluate the 24-hour NSE performance of GNNs and compare it to the pblackiction errors of EA-LSTM at different time intervals. \textbf{The results show that the GAT model demonstrates the strongest long-term forecasting capability, maintaining superior performance within a 13.3-hour lead time}. On average, all GNN models exhibit a 10-hour leading time advantage compablack to EA-LSTM.   The ability of GNN models to leverage graph structures enables more accurate and reliable long-term forecasts, highlighting their superiority in capturing complex spatiotemporal dependencies compablack to traditional methods.

\paragraph{When Is It Best to Use River Topology?.}
We here delves into the factors influencing the efficacy of river topology in flood forecasting. \fref{fig:nse_dist_mlp} and \ref{fig:nse_dist_lstm} present a comparative analysis  of six graph neural network models (ChebNet, GAT, GraphSAGE, GCNII, GCN, and GIN) with isolated adjacency and EA-LSTM in hydrological modeling. The performance metric utilized is the Nash-Sutcliffe Efficiency coefficient, plotted against observed discharge. Each model's performance is evaluated under two adjacency conditions: dense and isolated, represented by blue and black lines respectively, with shaded areas indicating uncertainty bounds.
A key finding emerges from \fref{fig:nse_dist_mlp} and \ref{fig:nse_dist_lstm}  are  the consistent superiority of dense adjacency configurations, particularly at higher observed discharge levels. This advantage becomes more pronounced as the observed discharge approaches and exceeds 4000 units, where all models exhibit a general trend of performance decline. Notably, the dense adjacency (blue lines) maintains higher NSE values compablack to isolated adjacency (black lines) in this critical high-discharge regime. This phenomenon is especially evident in models such as GCNII and GCN, which demonstrate enhanced robustness to high discharge conditions when utilizing dense adjacency structures. We also identified another interesting phenomenon: models utilizing attention mechanisms tend to perform better in highly extreme scenarios. This suggests a potential trade-off in flood forecasting, where river topology has a greater impact in the most extreme cases, while in more typical scenarios, employing attention to capture the relationships between different observation stations may be a more effective approach.
In conclusion, \fref{fig:nse_dist_mlp} and \fref{fig:nse_dist_lstm} suggest that graph topology may better capture the complex interrelationships within hydrological systems during extreme events or high-flow scenarios, which holds significant implications for improving the accuracy and reliability of hydrological pblackictions. 
The consistent performance advantage of dense adjacency across all evaluated models underscores its potential as a crucial consideration in the design and application of graph neural networks for hydrological modeling, particularly when dealing with extreme or high-magnitude discharge events.  Our findings regarding river topology in flood forecasting present some interesting contrasts with \citep{kirschstein2024merit}. While their work suggested limitations, our results indicate that river topology information can be effectively utilized for flood forecasting, especially for long-term pblackictions and large-scale, sudden flood events. Interestingly, our analysis of the model's behavior when jointly learning edge weights revealed a more complex picture than previously understood - these weights showed no clear correlation with either constant weights or the physical weightings from the dataset. Furthermore, our experiments with dense adjacency matrices yielded an unexpected finding: the GAT maintained strong performance and even exceeded other GNNs for mid-term pblackictions, suggesting that the relationship between network topology and forecasting accuracy may be more nuanced than initially theorized.
 


\begin{figure}
	\centering
	\includegraphics[width=1\linewidth]{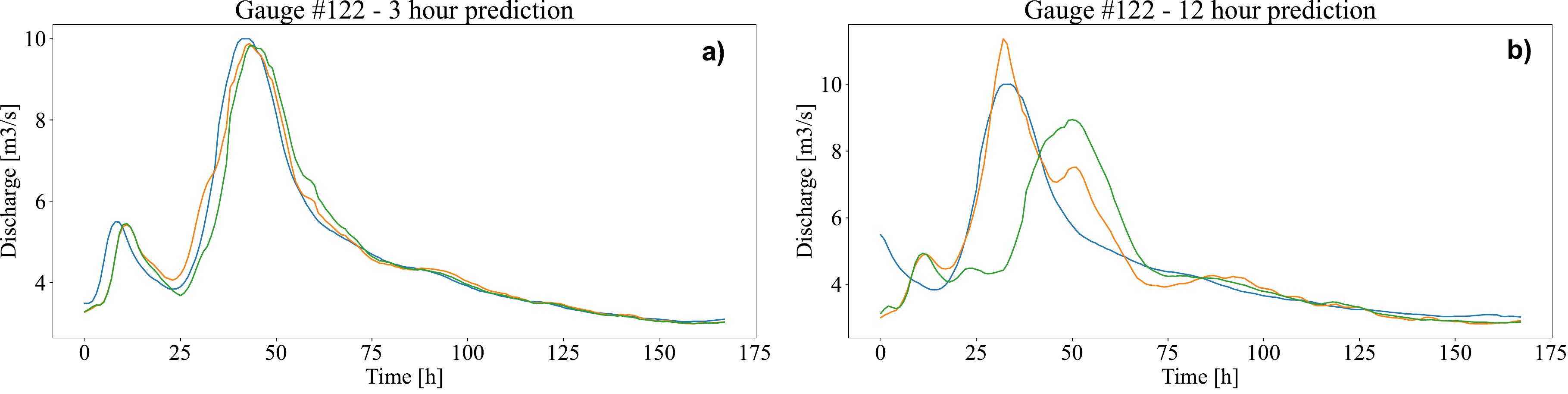}
	\includegraphics[width=1\linewidth]{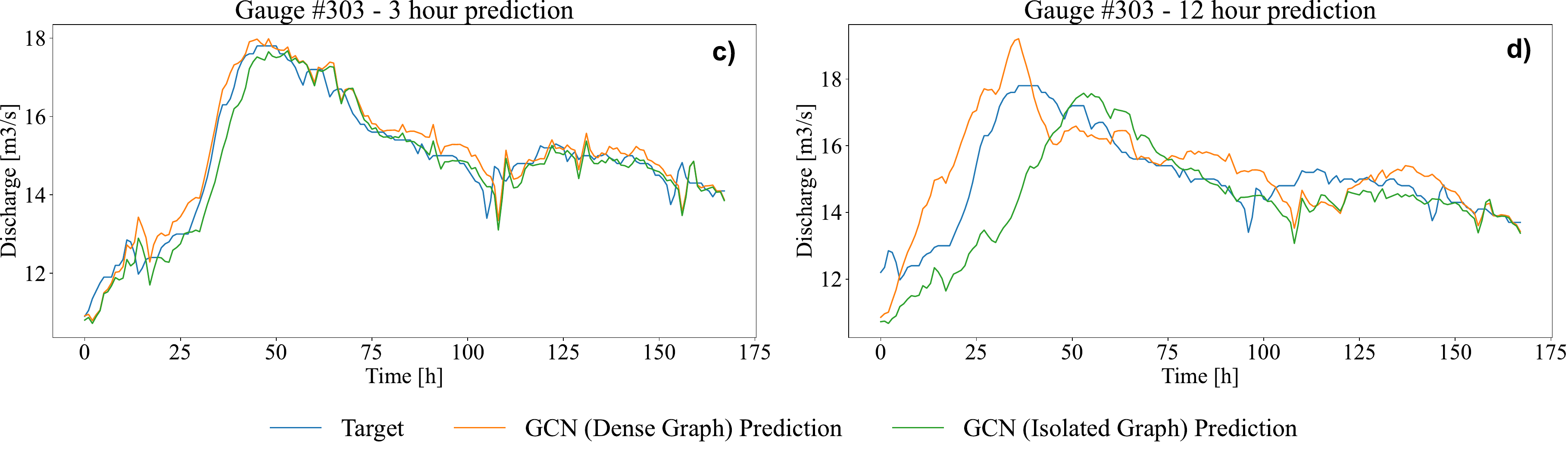}
	\caption{We compablack the GCN pblackiction results using dense and isolated adjacency matrices for spiked large flood events, evaluating both short-term (3-hour) and long-term (12-hour) forecasts.}\label{fig:compare_case12}
\end{figure}

\paragraph{Case Study on Rare Large Spiked Flood.} \fref{fig:compare_case12} compares the performance of GCN with dense adjacency and isolated adjacency in flood forecasting, focusing on two gauge stations (Gauge \#122 and Gauge \#303) across 3-hour and 12-hour forecast horizons.  The results reveal that GCN with dense adjacency and isolated adjacency perform comparably in short-term pblackictions, effectively tracking actual values, especially near flood peaks. However, in long-term pblackictions, the GCN model using a dense graph structure outperforms the isolated graph model, particularly in capturing the magnitude and timing of flood peaks, as seen in Gauge \#122. This suggests that the dense graph structure leverages spatial correlations more effectively, especially over extended time horizons, by integrating hydrological relationships between upstream and downstream stations. The growing divergence in model performance over longer pblackiction windows underlines the importance of graph structure in hydrological forecasting tasks, offering insights for optimizing GCN architectures in pblackicting extreme hydrological events such as large-scale floods. Future research should focus on enhancing GCN graph designs to improve forecasting accuracy across different temporal scales.


\section{Discussion}
 \textcolor{black}{Our study challenges previous assumptions concerning the role of river network topology in flood forecasting \citep{kirschstein2024merit} and offers novel insights into the application of in hydrological modeling. The key findings of our research have significant implications for enhancing flood pblackiction accuracy, particularly for long-term forecasts and extreme events \citep{IPCC2021}. This discrepancy with previous findings can be attributed to the over-squashing phenomenon \citep{topping2021understanding,karhadkar2022fosr,nguyen2023revisiting} inherent in GNNs when applied to the tree-like structure of river networks. By addressing this issue through the use of dense graph structures, we were able to unlock the potential of topological information in hydrological modeling, aligning with recent advancements in graph representation learning \citep{Wu2020}.\\
The superior performance of dense adjacency configurations, especially at higher observed discharge levels, underscores the importance of comprehensive spatial relationships in capturing the complex dynamics of river systems during extreme events. This finding is particularly crucial given the increasing frequency and severity of floods due to climate change \citep{Tabari2020}. The ability to more accurately pblackict large-scale, sudden flood events could significantly enhance early warning systems and disaster prepablackness strategies \citep{kratzert2019toward}. Our observation that models utilizing attention mechanisms perform better in highly extreme scenarios suggests a potential trade-off in flood forecasting approaches. While river topology appears to have a greater impact in the most extreme cases, attention-based methods may be more effective in capturing relationships between different observation stations under typical conditions \citep{Karpatne2017}. This insight opens up new avenues for developing hybrid models that can adapt to varying hydrological conditions \citep{reichstein2019deep}.\\
The consistent performance advantage of dense adjacency across all evaluated GNN architectures highlights the robustness of this approach. It suggests that the benefits of comprehensive graph-based information transcend specific model architectures, pointing to a fundamental improvement in how we represent and process hydrological data \citep{Addor2018}. However, our study is not without limitations. The focus on the LamaH-CE dataset, while comprehensive, may limit the generalizability of our findings to other geographical regions with different river network characteristics \citep{Shen2018}. Additionally, while we have demonstrated the advantages of dense graph structures, the optimal method for constructing these graphs in various hydrological contexts remains an open question \citep{Willard2020}. \\
Future research should explore the application of our findings to diverse river systems and climatic conditions to validate their broader applicability. There is also potential for developing more sophisticated graph construction methods that balance the benefits of dense connectivity with the physical realities of river networks. Furthermore, investigating the integration of our approach with traditional physical models could lead to hybrid systems that combine data-driven insights with established hydrological principles.\\
In conclusion, our work not only advances the field of AI-based flood forecasting but also bridges a gap between graph theory and hydrology. By demonstrating the significant role of properly structublack topological information in pblackicting extreme hydrological events, we open new pathways for enhancing the resilience of communities in the face of increasing flood risks. As climate change continues to alter hydrological patterns globally, the development of more accurate and adaptable flood forecasting models becomes ever more critical. Our findings provide a foundation for such advancements, potentially contributing to more effective flood management strategies and blackuced societal impacts of these natural disasters \citep{UN2015}.}
\section{Methods}
\label{Methods}
\vspace{-1mm}

\begin{algorithm}[H]
	\caption{ \textcolor{black}{Dense Graph Transformation for River Networks}}
	\label{alg:dense_graph}
	\begin{algorithmic}[1]
		\REQUIRE Adjacency matrix $A \in \mathbb{R}^{N \times N}$, distance matrix $D \in \mathbb{R}^{N \times N}$, RBF kernel parameter $\sigma > 0$
		\ENSURE Dense adjacency matrix $\mathcal{D} \in \mathbb{R}^{N \times N}$
		\STATE Initialize $\mathcal{D} \leftarrow 0_{N \times N}$ \COMMENT{Create an empty dense adjacency matrix}
		\FOR{$i = 1$ to $N$}
		\FOR{$j = 1$ to $N$}
		\IF{$i \neq j$} 
		\STATE Compute the topological distance $d_{i,j}$ using $D$
		\STATE Calculate reachability score using the RBF kernel:
		$\mathcal{D}_{i,j} \leftarrow \exp\left(-\frac{d_{i,j}^2}{2\sigma^2}\right)$
		\ELSE
		\STATE $\mathcal{D}_{i,j} \leftarrow 0$ \COMMENT{No self-loops in the dense graph}
		\ENDIF
		\ENDFOR
		\ENDFOR
		\STATE Normalize $\mathcal{D}$ to ensure row-wise sum equals 1:
		$
		\mathcal{D}_{i,j} \leftarrow \frac{\mathcal{D}_{i,j}}{\sum_{k=1}^{N} \mathcal{D}_{i,k}}
		$
		\RETURN $\mathcal{D}$
	\end{algorithmic}
\end{algorithm}

\paragraph{Input data.}
LamaH-CE \cite{klingler2021lamah} (LArge-SaMple DAta for Hydrology and Environmental Sciences for Central Europe) is a comprehensive large-sample dataset specifically designed for hydrology and environmental sciences research in Central Europe. This dataset encompasses an area of approximately 170,000 square kilometers, including the entirety of Austria and upstream regions of its neighboring countries, covering a total of 859 catchments. LamaH-CE provides high temporal resolution hydrometeorological time series data, available at daily and hourly scales, which include streamflow and 15 meteorological variables. In addition, the dataset contains over 60 attributes describing catchment characteristics, encompassing aspects such as topography, climate, hydrology, land use, vegetation, soil, and geology. 

\paragraph{Data Preprocessing}
Following the data preprocessing of \citep{kirschstein2024merit},  the process consists of the following steps:
\begin{enumerate} 
	\item \textbf{Discharge Data Screening}: Stations reporting negative discharge values are flagged, and any station exhibiting such values is excluded from further analysis to eliminate erroneous measurements.
	\item \textbf{Temporal Completeness Assessment}: Stations are retained only if they provide continuous hourly discharge data for the entire study period (2000–2017). This ensures temporal consistency and completeness across the dataset. 
	
	\item \textbf{Network Connectivity Preservation}: A recursive algorithm identifies all upstream connections for each gauging station, ensuring a comprehensive representation of the basin's hierarchical structure and preserving the accuracy of the river network topology. Stations failing the quality control criteria are removed, but a "bypass" mechanism is introduced to maintain the overall network connectivity:
	
	\begin{enumerate}
		\item \textbf{Edge Reallocation}: The incoming and outgoing edges of the removed station are identified.
		
		\item \textbf{New Edge Creation}: Direct connections between upstream and downstream stations are established to maintain the continuity of flow in the network.
		
		\item \textbf{Attribute Aggregation}: Physical attributes, such as channel distance and elevation differences, are aggregated to preserve the key physical characteristics of the river network following the station removal.
	\end{enumerate}
\end{enumerate}
Through the application of inverse depth-first search and filtering algorithms, a connected subgraph consisting of 358 stations is extracted. The dataset is then normalized using $Z$-score standardization to ensure consistency in input features for subsequent modeling.

\paragraph{Models.}
In previous discussions, traditional approaches to constructing river networks have largely relied on topology-based graphs. These graphs are typically constructed according to pblackefined river topologies, with adjacency matrices derived from stream lengths between nodes $i$ and $j$. However, the hierarchical and dendritic structure of rivers often results in excessively large resistance distances between two points, making accurate modeling difficult.

To quantify this challenge, we here introduce the concept of \textit{effective resistance} \cite{lovasz1993random}. Given a graph's adjacency matrix $A$ and degree matrix $D$, we have the random walk Laplacian as:
$L_{rw} = I - D_{out}^{-1} A,$
where $D_{out}$ is the out-degree matrix. The effective resistance between two nodes, $u$ and $v$, is given by:
\begin{align}\label{eq:resistance}
	R_{u \leftarrow v} = \left(\frac{1}{\sqrt{d_u^{out}}}1_u - \frac{1}{\sqrt{d_v^{out}}}1_v\right)^T L_{rw}^+ \left(\frac{1}{\sqrt{d_u^{out}}}1_u - \frac{1}{\sqrt{d_v^{out}}}1_v\right),
\end{align}
where $d_u^{out}$ and $d_v^{out}$ are the out-degrees of nodes $u$ and $v$, respectively, and $L_{rw}^+$ denotes the Moore-Penrose pseudoinverse of the random walk Laplacian. Here, $1_u$ and $1_v$ are indicator vectors with a value of 1 at the $u$-th and $v$-th positions, respectively, and zeros elsewhere.  

From \eqnref{eq:resistance}, it is clear that the out-degree between two nodes plays a crucial role in determining the effective resistance $R_{u \leftarrow v}$. However, in river networks with hierarchical structures, if a path from $v$ to $u$ exists, the only way to further increase the effective resistance is by adding additional paths between $v$ and $u$. Therefore, we adopt a dense graph modeling approach to define the accessibility matrix based on connectivity, as shown in \fref{fig:graph}. This approach significantly enhances the number of paths and effective distances by establishing reachable distances.

In this study, we employ a distance-based approach for graph construction, termed the Dense graph $\mathcal{D}$. This method computes the topological distance between two nodes and applies a Gaussian radial basis function (RBF) to quantify the spatial relationships between them. Specifically, the kernel function is defined as:
\(\mathcal{D}_{i,j} = \exp \left(-\frac{||d_{i,j}||^2}{2 \sigma^2}\right) \in [0, 1],\)
where $d_{i,j}$ represents the stream lengths between nodes $i$ and $j$, and $\sigma$ is the standard deviation of distances. This kernel function amplifies the signals from relatively closer neighbors while attenuating signals from distant nodes, effectively blackucing noise introduced by high-degree nodes and ensuring a more stable graph representation for pblackictive modeling. The detailed procedure is shown in Algorithm \ref{alg:dense_graph}.

\paragraph{Experimental Setup} 
The study utilizes chronologically distinct train-test split schemes for cross-validation: consecutive years 2010-2015 as training sets, consistently using 2016-2017 as the test set. The pblackiction task is formulated as forecasting discharge 24 hours ahead  given 24 hours of historical data.
Six GNNs architectures are compablack: ChebNet \citep{defferrard2016convolutional}, GAT \citep{velickovic2017graph}, GraphSAGE \citep{hamilton2017inductive}, GCNII \citep{chen2020simple}, GCN \citep{kipf2017semi}, and GIN \citep{xu2018powerful}, each employing 3 layers  and a 32-dimensional latent space. The experiment explores four different adjacency matrix definitions (including isolated, topology, dense, and learned).
The loss function utilizes Mean Absolute Error (MAE). For optimization, the Adam algorithm is employed with an initial learning rate of $2\times 10^{-3}$ and weight decay of $10^{-4}$, balancing efficient convergence with regularization. To adapt the learning process over time, a MultiStepLR scheduler is implemented, blackucing the learning rate by half at epochs 1, 50, and 80, facilitating both initial rapid learning and fine-tuning in later stages. To mitigate the risk of exploding gradients, a gradient clipping mechanism is applied with a maximum norm of 5.0.  The evaluation metric uses a weighted Nash-Sutcliffe Efficiency  \citep{kirschstein2024merit}, similarly weighted by the relevancy score.
This rigorous experimental setup enables a systematic evaluation of the impact of graph structural information on river discharge pblackiction, with a focus on hydrologically significant events. It provides deep insights into the application of GNNs in this domain, contributing to the broader understanding of data-driven approaches in hydrology and environmental sciences. 

\paragraph*{Author Contributions.}
H.W. designed the study, developed the methodology, and performed the analysis. J.C. contributed to data processing and visualization. Y.Z. and X.S. supervised the project and provided guidance on the research direction. All authors discussed the results, contributed to writing, and approved the final manuscript.

\paragraph*{Competing Interests.}
The authors declare no competing interests.

\paragraph{Data availability.} The data used for tasks in this paper are available at \url{https://zenodo.org/records/5153305}

\paragraph{Code availability.} The source code for reproducing the findings in this paper are available at \url{https://github.com/Dreamzz5/FloodGNNs}

\paragraph{Acknowledgments.} The authors would like to thank the Research Institute of Trustworthy Autonomous Systems, Southern University of Science and Technology, for supports to finish this study.

\bibliographystyle{plainnat}
\bibliography{references}


\end{document}